\documentclass{article}

\usepackage{arxiv}

\usepackage[utf8]{inputenc} 
\usepackage[T1]{fontenc}    
\usepackage{hyperref}       
\usepackage{url}            
\usepackage{booktabs}       
\usepackage{amsfonts}       
\usepackage{nicefrac}       
\usepackage{microtype}      
\usepackage{lipsum}		
\usepackage{graphicx}
\usepackage{natbib}
\usepackage{doi}

\usepackage{soul}
\usepackage[utf8]{inputenc}
\usepackage{graphicx}
\usepackage{amsmath}
\usepackage{amsthm}
\usepackage{booktabs}

\usepackage{epsfig}

\usepackage{amssymb}
\usepackage{epsfig}
\usepackage{mathtools}
\usepackage[boxruled]{algorithm2e}

\DeclareMathOperator*{\argmin}{\arg\!\min}
\usepackage{multirow}
\usepackage{booktabs}
\usepackage{balance}
\usepackage{subfig}

\newtheorem{theorem}{Theorem}
\usepackage[colorinlistoftodos]{todonotes}

\title{Multiscale Manifold Warping}
\author{

       Sridhar Mahadevan \And 
       Anup Rao \And 
       Georgios Theocharous \And   Jennifer Healey\\
   Adobe Research, 345 Park Avenue, San Jose, CA 95110 \\
   
   \{smahadev, anuprao, theochar, jehealey\}@adobe.com
}

\begin{document}
\maketitle

\begin{abstract}
Many real-world applications require aligning two temporal sequences, including bioinformatics, handwriting recognition, activity recognition, and human-robot coordination. Dynamic Time Warping (DTW) is a popular alignment method, but can fail on high-dimensional real-world data where the dimensions of aligned sequences are often unequal. In this paper, we show that exploiting the multiscale manifold latent structure of real-world data can yield improved alignment. We introduce a novel framework called Warping on Wavelets (WOW) that integrates DTW with a a multi-scale manifold learning framework called Diffusion Wavelets.  We present a theoretical analysis of the WOW family of algorithms and show that it outperforms previous state of the art methods, such as canonical time warping (CTW) and manifold warping, on several real-world datasets.
\end{abstract}

\section{Introduction}
\label{intro}

\begin{figure}  [h]
  \centering
    \includegraphics[width=0.5\columnwidth]{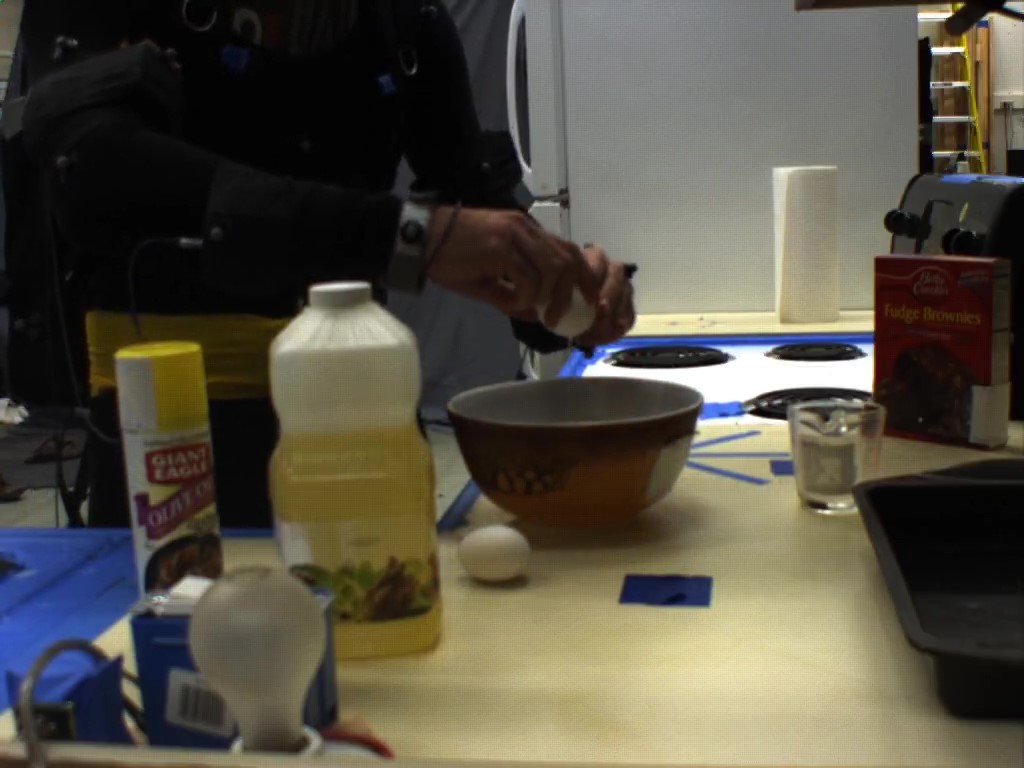}
     \includegraphics[width=0.5\columnwidth]{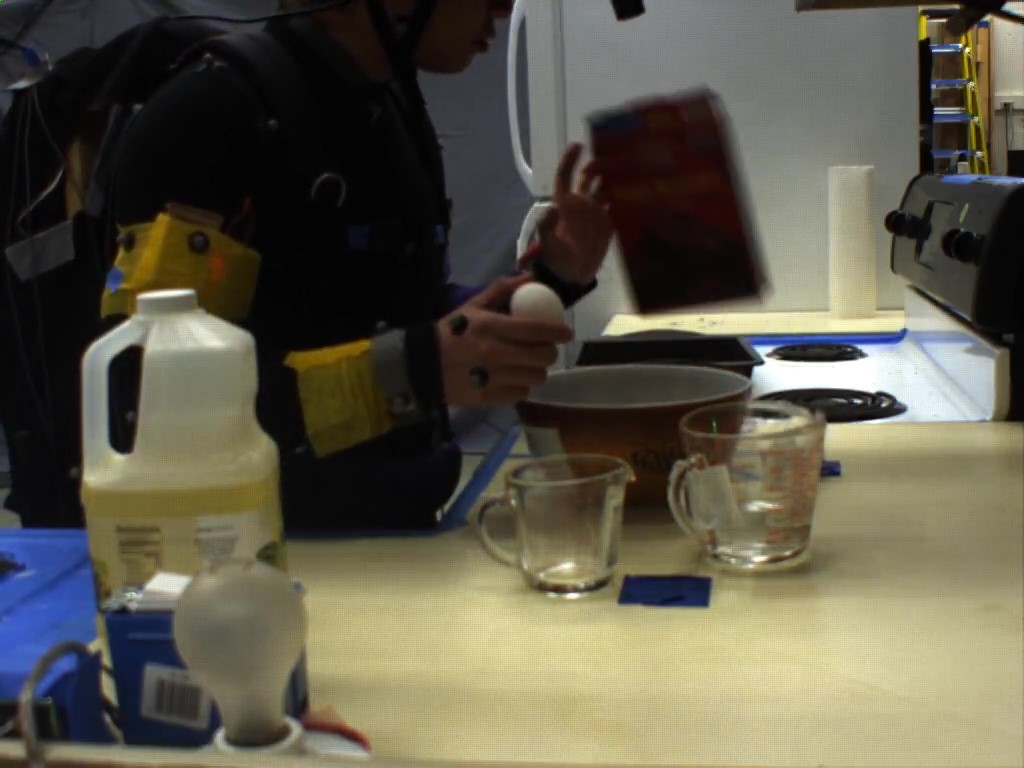}
  \caption{A real-word  problem of aligning human motion from multimodal capture data \citep{torre2008motiondata} from the CMU Quality of Life Grand Challenge, which records human subjects cooking a variety of dishes. We propose a novel multiscale manifold framework to align such high dimensional time-series data.}
  \label{fig:dtw}
\end{figure}

Temporal alignment of time series is central to many real-world applications, including human motion recognition (see Figure~\ref{fig:dtw}) \citep{junejo2008activity_alignment}, temporal segmentation \citep{zhou2008aligned},  modeling the spread of  Covid-19 \citep{covid-dtw}, and building view-invariant representations of activities \citep{junejo2008activity_alignment}.
Dynamic time warping (DTW) \citep{sakoe1978dtw} is a widely-used classical approach to aligning time-series datasets. DTW requires an inter-set distance function, and often assumes both input data sets have the same dimensionality. DTW may also fail under arbitrary affine transformations of one or both inputs.
Canonical time warping (CTW) \citep{zhou2009ctw} combines DTW by with canonical correlation analysis (CCA) \citep{anderson2003cca}  to find a joint lower-dimensional embedding of two time-series datasets, and subsequently align the datasets in the lower-dimensional space. However, CTW fails when the two related data sets require nonlinear transformations.  Manifold warping \citep{hoa-cj-mw}\citep{ham2003MA,wang2009generalMAframework} solved this by instead representing features in the latent joint manifold space of the sequences. 

Prior manifold warping methods however do not exploit the multiscale nature of most datasets, which our proposed algorithms exploit. In this paper, we propose
a novel variant of dynamic time warping 
that uses a type of multiscale {\em wavelet} analysis
\citep{mallat} on graphs, called {\em diffusion wavelets} \citep{dwt}
to address this gap. In particular, we develop a multiscale variant of manifold warping called WOW (warping on wavelets), and show that WOW outperforms several warping algorithms, including manifold warping, as well as two other novel warping methods.

\section{Dynamic Time Warping}
\label{dtw}

We give a brief review of dynamic time warping \citep{sakoe1978dtw}. We are given two sequential data sets $X= [ x_{1}^{T},\ldots,x_{n}^{T} ]^{T} \in \mathbb{R}^{n \times d} $, $Y=[ y_{1}^{T},\ldots,y_{m}^{T}]^{T}  \in \mathbb{R}^{m \times d}$ in the same space with a distance function $dist:X \times Y \rightarrow \mathbb{R}$. 
Let $P=\{p_1,...,p_s\}$ represent an alignment between $X$ and $Y$,
where each $p_k=(i,j)$ is a pair of indices such that $x_i$ corresponds with $y_j$.
Since the alignment is restricted to sequentially-ordered data, we impose the additional constraints:
\begin{eqnarray}
p_{1} & = & (1,1)  \label{eq:Wcontraint1} \\
p_{s} & = & (n,m)  \label{eq:Wcontraint2} \\
p_{k+1}-p_{k} & = & (1,0)\: or\:(0,1)\: or\:(1,1) \label{eq:Wcontraint3}
\end{eqnarray}
A valid alignment must match the first and last instances and cannot skip any intermediate instance.
Also, no two sub-alignments cross each other. We can also represent the alignment in matrix form $W$ where:
\begin{equation}
	W_{i,j}=  \left\{  \begin{array}{ll} 
	1 & \mbox{if $(i,j) \in P$} \\ 
	0 & \mbox{otherwise} \end{array}   \right.  
\end{equation}
To ensure that $W$ represents an alignment which satisfies the constraints in Equations \ref{eq:Wcontraint1}, \ref{eq:Wcontraint2}, \ref{eq:Wcontraint3}, $W$ must be in the following form:
$W_{1,1}=1,W_{n,m}=1$, none of the columns or rows of $W$ is a $0$ vector, and there must not be any $0$ between any two $1$'s in a row or column of $W$.
We call a $W$ which satifies these conditions a \emph{DTW matrix}.
An optimal alignment is the one which minimizes the loss function with respect to the DTW matrix $W$:
\begin{eqnarray}
	L_{\mbox{DTW}}(W) = \sum_{i,j}dist\left(x_i,y_j\right)W_{i,j}
\end{eqnarray}
A na\"{\i}ve search over the space of all valid alignments would take exponential time;
however, dynamic programming can produce an optimal alignment in $O(nm)$. When $m$ is high-dimensional, as in Figure~\ref{fig:dtw}, or if the two sequences have varying dimensionality, DTW is not as effective, and we turn next to discussing a broad framework  to extend DTW based on exploiting the manifold nature of many real-world datasets.

\section{Mutiscale Manifold Learning}
\label{dwt} 

Diffusion wavelets (DWT) \citep{dwt} extends
the strengths of classical wavelets to data that lie on graphs and
manifolds. The term {\em diffusion wavelets} is used because it is
associated with a diffusion process that defines the different
scales, allows a multiscale analysis of functions on manifolds and
graphs. 

The diffusion wavelet procedure is described in
Figure~\ref{fig:dwt}. The main procedure is as
follows: an input matrix $T$ is orthogonalized using an
approximate $QR$ decomposition in the first step. $T$'s $QR$
decomposition is written as $T= QR$, where $Q$ is an orthogonal
matrix and $R$ is an upper triangular matrix. The orthogonal
columns of $Q$ are the scaling functions. They span the column
space of matrix $T$. The upper triangular matrix $R$ is the
representation of $T$ on the basis $Q$. In the second step, we
compute $T^2$. Note this is not done simply by multiplying $T$ by
itself. Rather, $T^2$ is represented on the new basis $Q$:
$T^2=(RQ)^2$.  Since $Q$ may have fewer columns than
$T$, due to the approximate QR decomposition, $T^2$ may be a smaller square matrix. The above process is repeated at the next level, generating compressed dyadic powers
$T^{2^j}$,  until the maximum level is reached or its effective
size is a $1 \times 1$ matrix. Small powers of $T$ correspond to
short-term behavior in the diffusion process and large powers
correspond to long-term behavior. 

\begin{figure}[p]
\center\footnotesize
\begin{tabular}{|p{12cm}|}
\hline\\ $\{\phi_j$, $T_j \}=DWT(T, \phi_0, QR, J, 
\varepsilon)$\\
$\,\,  \textbf{INPUT}$:\\
$\,\,T$: Diffusion operator. \\  $\phi_0$: Initial basis matrix. \\ $QR$: A modified $QR$ decomposition.\\
$J$: Max step number\\
$\varepsilon$: Desired precision.\\
$\,\, // \textbf{OUTPUT}: \phi_j$: Diffusion scaling functions at scale $j$. $T_j=[T^{2^j}]_{\phi_j}^{\phi_j}$.\\
$For \,\,j=0\,\,\, to \,\,\, J-1$
$\{$\\
$\,\,\,\,\,\,\,\,([\phi_{j+1}]_{\phi_j}$, $[T^{2^j}]_{\phi_j}^{\phi_{j+1}}) \leftarrow QR([T^{2^j}]_{\phi_j}^{\phi_j}, \varepsilon)$;\\
$\,\,\,\,\,\,\,\,[T^{2^{j+1}}]_{\phi_{j+1}}^{\phi_{j+1}}=([T^{2^j}]_{\phi_j}^{\phi_{j+1}} [\phi_{j+1}]_{\phi_j})^2$;\\
$\}$\\ \hline
\end{tabular}
\caption{\small Diffusion Wavelets construct multiscale
representations at different scales. The notation
$[T]_{\phi_a}^{\phi_b}$ denotes matrix $T$ whose column space is
represented using basis $\phi_b$ at scale $b$, and row space is
represented using basis $\phi_a$ at scale $a$. The notation
$[{\phi_b}]_{\phi_a}$ denotes basis $\phi_b$ represented on the
basis $\phi_a$. At an arbitrary scale $j$, we have $p_j$ basis
functions, and length of each function is $l_j$.
$[T]_{\phi_a}^{\phi_b}$ is a $p_b \times l_a$ matrix,
$[{\phi_b}]_{\phi_a}$ is an $l_a \times p_b$ matrix.}\label{fig:dwt}
\end{figure}

\begin{figure}[p]
  \centering
    \includegraphics[height = 12cm, width=0.45\columnwidth,keepaspectratio]{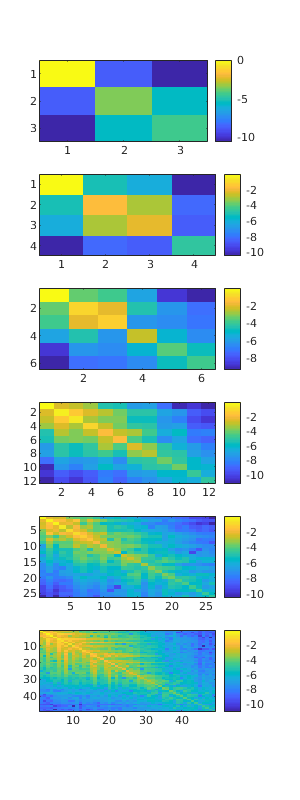}
  \caption{The diffusion wavelet procedure on a real-world CMU Quality of Life dataset (see Figure~\ref{fig:dtw}), where a subject is making brownies. The two sensor streams being aligned are an $87$-dimensional motion capture stream and a $11$-dimensional internal measurement unit system.  The  figure shows the diffusion wavelet tree constructed. The diffusion operator at each level is represented on the basis constructed at the previous level. The color values are scaled logarithmically.}
  \label{fig:dwt-quality-of-life}
\end{figure}

An example of multiscale tree constructed by the diffusion wavelet procedure is shown in Figure~\ref{fig:dwt-quality-of-life}, which is one of the real-world domains that we study later in the paper. 

We introduce multiscale  Laplacian
eigenmaps~\citep{belkin01} and locality preserving projections (LPP)~\citep{he-lpp-03}. Laplacian
eigenmaps construct embeddings of data using the low-order
eigenvectors of the graph Laplacian as a 
basis~\citep{chung-laplacian-97}, which extends Fourier analysis to
graphs and manifolds. Locality Preserving Projections (LPP)  is a
linear approximation of Laplacian eigenmaps. We review the multiscale Laplacian eigenmaps and multiscale LPP, based on the diffusion wavelets framework \citep{chang-aaai2013}. 

{\bf Notation:} $X= [x_1, \cdots, x_n]$ be an $p \times n$ matrix
representing $n$ instances defined in a $p$ dimensional space. $W$
is an $n \times n$ weight matrix, where $W_{i,j}$ represents the
similarity of $x_i$ and $x_j$ ($W_{i,j}$ can be defined by
$e^{-\|x_i-x_j\|^2}$). $D$ is a diagonal valency matrix, where
$D_{i,i}=\sum_j W_{i,j}$. $\mathcal{W}=D^{-0.5}WD^{-0.5}$.
$\mathcal{L}= I- \mathcal{W}$, where $\mathcal{L}$ is the
normalized Laplacian matrix and $I$ is an identity matrix. $XX^T=
FF^T$, where $F$ is a $p \times r$ matrix of rank $r$. One way to
compute $F$ from $X$ is singular value decomposition. $(\cdot)^+$
represents the Moore-Penrose pseudo inverse.
\\\textbf{(1) Laplacian
eigenmaps} minimizes the cost function
\textbf{$\sum_{i,j}(y_i-y_j)^2\mathcal{W}_{i,j}$}, which
encourages the neighbors in the original space to be neighbors in
the new space.  The $c$ dimensional embedding is provided by
eigenvectors of $\mathcal{L}x= \lambda x$ corresponding to the $c$
smallest non-zero eigenvalues.   The cost function for
\emph{\textbf{multiscale Laplacian eigenmaps}} is defined as
follows: given $X$, compute $Y_k=[y_k^1, \cdots, y_k^n]$ at level
$k$ ($Y_k$ is a $p_k \times n$ matrix) to minimize
$\sum_{i,j}(y_k^i-y_k^j)^2\mathcal{W}_{i,j}$.
Here $k=1, \cdots,
J$ represents each level of the underlying manifold hierarchy.
\\\textbf{(2) LPP} is a linear approximation of Laplacian eigenmaps.
LPP minimizes the cost function
$\sum_{i,j}(f^Tx_i-f^Tx_j)^2\mathcal{W}_{i,j}$, where the mapping
function $f$  constructs a $c$ dimensional embedding, and is
defined by the eigenvectors of $X\mathcal{L}X^Tx= \lambda XX^Tx$
corresponding to the $c$ smallest non-zero eigenvalues. Similar to
multiscale Laplacian eigenmaps, \emph{\textbf{multiscale LPP}}
learns linear mapping functions defined at multiple scales to
achieve multilevel decompositions.

\subsection{The Multiscale Algorithms}

\begin{figure}[t]
\center\footnotesize
\begin{tabular}{|p{12cm}|}
\hline

\begin{enumerate}

\item {\bf Construct diffusion matrix $T$ characterizing the given
data set:}
\begin{itemize}
\item {$T= I-\mathcal{L}$ \text{is an $n \times n$ diffusion matrix.}}
\end{itemize}

\item {\bf Construct multiscale basis functions using diffusion
wavelets:}

\begin{itemize}
\item $\{\phi_j, T_j\}=DWT(T, I, QR, J, \varepsilon)$.

\item {The resulting $[\phi_j]_{\phi_0}$ is an $n \times p_j$
matrix (Equation (\ref{equ:dp})).}

\end{itemize}

\item {\bf Compute lower dimensional embedding (at level $j$):}
\begin{itemize}
\item {The embedding $x_i \rightarrow y_i=$ row $i$ of
$[\phi_j]_{\phi_0}$.}

\end{itemize}
\end{enumerate}
\\\hline
\end{tabular}

\vskip 0.05in
\begin{tabular}{|p{12cm}|}
\hline

\begin{enumerate}

\item {\bf Construct relationship matrix $T$ characterizing the
given data set:}
\begin{itemize}
\item {$T= (F^+X\mathcal{L}X^T(F^T)^+)^+$ \text{is an $r \times r$
matrix.}.}
\end{itemize}

\item {\bf Apply diffusion wavelets to explore the intrinsic
structure of the data:}
\begin{itemize}
\item $\{\phi_j, T_j\}= DWT(T, I, QR, J, \varepsilon)$.
\end{itemize}
\begin{itemize}
\item {The resulting $[\phi_j]_{\phi_0}$ is an $r \times p_j$
matrix (Equation (\ref{equ:dp})).}
\end{itemize}

\item {\bf Compute lower dimensional embedding (at level $j$):}
\begin{itemize}

\item { The embedding $x_i \rightarrow y_i=
{((F^T)^+[\phi_j]_{\phi_0})}^T x_i$.}
\end{itemize}
\end{enumerate}
\\\hline
\end{tabular}
\caption{Top: Multiscale Laplacian Eigenmaps; Bottom: Multiscale
LPP.}\label{fig:alg}
\end{figure}

Multiscale Laplacian eigenmaps and multiscale LPP algorithms are
shown in Figure~\ref{fig:alg}, where $[\phi_j]_{\phi_0}$ is used
to compute a lower dimensional embedding. As shown in
Figure~\ref{fig:dwt}, the scaling functions
$[\phi_{j+1}]_{\phi_j}$ are the orthonormal bases 
that
span the column space of $T$ at different levels. They define a set of new
coordinate systems revealing the information in the original
system at different scales. The scaling functions also provide a
mapping between the data at longer spatial/temporal scales and
smaller scales. Using the scaling functions, the basis functions
at level $j$ can be represented in terms of the basis functions at
the next lower level. In this manner, the extended basis functions
can be expressed in terms of the basis functions at the finest
scale using: {\small
\begin{equation}\label{equ:dp}
[\phi_j]_{\phi_0}= [\phi_j]_{\phi_{j-1}}[\phi_{j-1}]_{\phi_0} =
[\phi_j]_{\phi_{j-1}}\cdots [\phi_1]_{\phi_0} [\phi_0]_{\phi_0},
\end{equation}}where each element on the right hand side of the equation is
created by the procedure shown in Figure~\ref{fig:dwt}. In our
approach, $[\phi_j]_{\phi_0}$ is used to compute lower dimensional
embeddings at multiple scales. Given $[\phi_j]_{\phi_0}$, any
vector/function on the compressed large scale space can be
extended naturally to the finest scale space or vice versa. The
connection between vector $v$ at the finest scale space and its
compressed representation at scale $j$ is computed using the
equation
 $[v]_{\phi_0}=([\phi_j]_{\phi_0})[v]_{\phi_j}$. The
elements in $[\phi_j]_{\phi_0}$ are usually much coarser and
smoother than the initial elements in $[\phi_0]_{\phi_0}$, which
is why they can be represented in a compressed form.

\section{Multiscale Manifold Alignment}
\label{mma} 

We describe a general framework for transfer learning across two datasets called manifold alignment \citep{yunqian11,wang2009generalMAframework}.
We are given the data sets $X$ and $Y$ of shapes $N_X \times D_X$ and $N_Y \times D_Y$,
where each row is a sample (or instance) and each column is a feature,
and a correspondence matrix $C^{(X,Y)}$ of shape $N_X\times N_Y$, where
\begin{equation}
  \label{wxy}
  C_{i,j}^{(X,Y)} = \left\{
     \begin{array}{ll}
       1&:\text{$X_i$ is in correspondence with $Y_j$}\\
       0&:\text{otherwise}
     \end{array}
   \right..
\end{equation}

Manifold alignment calculates the embedded matrices $F^{(X)}$ and $F^{(Y)}$ of shapes $N_X \times d$ and $N_Y \times d$ for $d\le min(D_X,D_Y)$ that are the embedded representation of $X$ and $Y$ in a shared, low-dimensional space.
These embeddings aim to preserve both the intrinsic geometry within each data set and the sample correspondences among the data sets.
More specifically, the embeddings minimize the following loss function: 
\begin{align}
  \label{man_loss1}
  L_{\mbox{MA}}\left(F^{(X)},F^{(Y)}\right) &= \frac{\mu}{2} \sum_{i=1}^{N_X}\sum_{j=1}^{N_Y} ||F_i^{(X)} - F_j^{(Y)}||_2^2 C_{i,j}^{(X,Y)}\nonumber\\
  &+ \frac{1-\mu}{2} \sum_{i,j=1}^{N_X} ||F_i^{(X)}-F_j^{(X)}||_2^2 W_{i,j}^{(X)}\nonumber\\
  &+ \frac{1-\mu}{2} \sum_{i,j=1}^{N_Y} ||F_i^{(Y)}-F_j^{(Y)}||_2^2 W_{i,j}^{(Y)},
\end{align}
where $N$ is the total number of samples $N_X+N_Y$,
$\mu\in[0,1]$ is the correspondence tuning parameter,
and $W^{(X)},W^{(Y)}$ are the calculated similarity matrices of shapes $N_X\times N_X$ and $N_Y\times N_Y$, such that
\begin{equation}
  W_{i,j}^{(X)} = \left\{
     \begin{array}{ll}
       k(X_i,X_j) &:\text{$X_j$ is a neighbor of $X_i$}\\
       0 &:\text{otherwise}
     \end{array}
   \right.
\end{equation}
for a given kernel function $k(\cdot,\cdot)$.
$W_{i,j}^{(Y)}$ is defined in the same fashion.
Typically, $k$ is set to be the nearest neighbor set member function or the heat kernel \\$k(X_i,X_j)=\exp{\left(-|X_i-X_j|^2\right)}$.

In the loss function of equation (\ref{man_loss1}),
the first term corresponds to the alignment error between corresponding samples in different data sets.
The second and third terms correspond to the local reconstruction error for the data sets $X$ and $Y$ respectively.
This equation can be simplified using block matrices by introducing a joint weight matrix $W$ and a joint embedding matrix $F$, where
\begin{equation}
  W = \left[
    \begin{array}{cc}
      (1-\mu)W^{(X)} & \mu C^{(X,Y)} \\
      \mu C^{(Y,X)} & (1-\mu)W^{(Y)}
    \end{array}
  \right]
\end{equation}
and
\begin{equation}
  F = \left[
    \begin{array}{c}
      F^{(X)}\\
      F^{(Y)}
    \end{array}
    \right].
\end{equation}

 \subsection{Multiscale alignment} 

Given a fixed sequence of dimensions, $d_1 > d_2 > \ldots > d_h$,
as well as two datasets, $X$ and $Y$, and some partial
correspondence information, $x_i \in X_l \longleftrightarrow y_i
\in Y_l$,the multiscale manifold alignment problem is to compute
mapping functions, $\mathcal{A}_k$ and $\mathcal{B}_k$, at each
level $k$ ($k = 1,2,\ldots,h$) that project $X$ and $Y$ to a new
space, preserving local geometry of each dataset and matching
instances in correspondence. Furthermore, the associated sequence
of mapping functions should satisfy $span(\mathcal{A}_1) \supseteq
span(\mathcal{A}_2) \supseteq \ldots \supseteq
span(\mathcal{A}_h)$ and $span(\mathcal{B}_1) \supseteq
span(\mathcal{B}_2) \supseteq \ldots \supseteq
span(\mathcal{B}_h)$, where $span(\mathcal{A}_i)$ (or
$span(\mathcal{B}_i)$)represents the subspace spanned by the
columns of $\mathcal{A}_i$ (or $\mathcal{B}_i$).

To apply diffusion wavelets to the multiscale alignment problem, the construction needs to be able to handle two input matrices $A$ and $B$ that occur in  a generalized eigenvalue decomposition, $A
\gamma= \lambda B \gamma$. The following theoretical result shows how to carry out such an extension \citep{wang13}. Given $X, X_l, Y, Y_l$, using the notation defined in
Figure~\ref{fig:notationmsma}, the algorithm is given below as Algorithm 2.

\begin{algorithm}[t]
\caption{Multiscale Manifold Alignment (MMA)}
\SetAlgoLined
\begin{enumerate}
\small \item {\bf Construct a matrix representing the joint
manifold:} $T=F^+ZLZ^T(F^T)^+$.

\item {\bf Use diffusion wavelets on the joint manifold:}

 $[\phi_k]_{\phi_0}= \mathcal{DWT}(T^+, \epsilon)$, where $\mathcal{DWT()}$
is the diffusion wavelets algorithm. 

\item {\bf Compute mapping functions for manifold alignment (at
level $k$):}\\
$\left[%
\begin{array}{c}
  \alpha_k\\
  \beta_k\\
\end{array}%
\right]=(F^T)^+[\phi_k]_{\phi_0}$ is a $(p+q) \times d_k$ matrix.

\item{\bf At level $k$: apply $\alpha_k$ and $\beta_k$ to find
correspondences between $X$ and $Y$:}

For any $i$ and $j$, $\alpha_k^Tx_i$ and $\beta_k^Ty_j$ are in the
same $d_k$ dimensional space.
\end{enumerate} 
\end{algorithm}



\begin{figure}[h]
\center
\begin{tabular}{|p{12cm}|}
\hline \vskip 0.01in $x_i \in R^p$; $X=\{x_1, \cdots, x_m\}$ is a
$p \times m$ matrix; \\ $X_l=\{x_1, \cdots, x_l\}$ is a $p \times l$
matrix.
\\$y_i \in R^q$; $Y=\{y_1, \cdots, y_n\}$ is a $q \times
n$ matrix;\\ $Y_l=\{y_1, \cdots, y_l\}$ is a $q \times l$ matrix .
\\$X_l$ and $Y_l$ are in correspondence: $x_i \in X_l \longleftrightarrow y_i \in Y_l$.
\\$W_x$ is a similarity matrix, e.g. $W_x^{i,j} = e^{-\frac{|| x_i - x_j||^2}{2\sigma^2}}$.
\\$D_x$ is a full rank diagonal matrix: $D_x^{i,i}=\sum_j W_x^{i,j}$;
\\$L_x= D_x- W_x$ is the combinatorial Laplacian matrix.
\\$W_y$, $D_y$ and $L_y$ are defined similarly.
\\\\$\Omega_1-\Omega_4$ are all diagonal matrices having $\mu$ on
the top $l$ elements \\ of the diagonal (the other elements are 0s);\\ 
$\Omega_1$ is an $m \times m$ matrix; $\Omega_2$ and $\Omega_3^T$
are $m \times n$ matrices; \\ $\Omega_4$ is an $n \times n$ matrix.
\\$Z=\left(%
\begin{array}{cc}
  X & 0 \\
  0 & Y \\
\end{array}%
\right)$ is a $(p+q) \times (m+n)$ matrix.\\
$D=\left(%
\begin{array}{cc}
  D_x & 0 \\
  0 & D_y \\
\end{array}%
\right)$ and $L=\left(%
\begin{array}{cc}
  L_x+\Omega_1 & -\Omega_2 \\
  -\Omega_3 & L_y+\Omega_4 \\
\end{array}%
\right)$ \\ are both $(m+n) \times (m+n)$ matrices.
\\$F$ is a $(p+q)\times r$ matrix, where $r$ is the rank of
$ZDZ^T$ \\ and $FF^T=ZDZ^T$. $F$ can be constructed by SVD.
\\$(\cdot)^{+}$ represents the Moore-Penrose pseudoinverse.
\\
\\At level $k$: $\alpha_k$ is a mapping from $x \in X$ to a point, \\ $\alpha_k^Tx$, in a $d_k$ dimensional space
($\alpha_k$ is a $p \times d_k$ matrix).
\\At level $k$: $\beta_k$ is a mapping from $y \in Y$ to a point, \\ $\beta_k^Ty$, in a $d_k$ dimensional space \\ ($\beta_k$ is a $q \times d_k$
matrix). 
\\\hline
\end{tabular}
\caption{Notation used in this section.}\label{fig:notationmsma}
\end{figure}

%


%
\begin{theorem}
The solution to the generalized eigenvalue
decomposition $ZLZ^T\gamma = \lambda ZDZ^T\gamma$ is given by
$((F^T)^+x, \lambda)$, where $x$ and $\lambda$ are eigenvector and
eigenvalue of $F^+ZLZ^T(F^T)^+x= \lambda
x$.\label{theorem:generalizedeigen}
\end{theorem}

{\bf Proof:}
Using the notation summarized in Figure~\ref{fig:notationmsma},
$ZDZ^T= FF^T$, where $F$ is a $(p+q) \times r$ matrix of rank $r$
and can be constructed by singular value decomposition. It is
obvious that $ZDZ^T$ is positive semi-definite.
\\\textbf{Case 1:} when $ZDZ^T$ is positive definite:
\\It can be seen that $r =p+q$. This implies that $F$ is a $(p+q) \times (p+q)$ full rank
matrix: $F^{-1}=F^+$.
\\$ZLZ^T\gamma = \lambda ZDZ^T\gamma$
$\Longrightarrow ZLZ^T\gamma= \lambda FF^T\gamma$
$\Longrightarrow ZLZ^T\gamma = \lambda F (F^T\gamma)$
\\$\Longrightarrow ZLZ^T(F^T)^{-1}(F^T\gamma)= \lambda F (F^T\gamma)$
$\Longrightarrow F^{-1}ZLZ^T(F^T)^{-1}(F^T\gamma)=  \lambda
(F^T\gamma)$
\\$\Longrightarrow$ Solution to $ZLZ^T\gamma = \lambda ZDZ^T\gamma$ is given by $((F^T)^+x,
\lambda)$, where $x$ and $\lambda$ are eigenvector and eigenvalue
of $F^+ZLZ^T(F^T)^+x=  \lambda x$.
\\\textbf{Case 2:} when $ZDZ^T$ is positive semi-definite but not
positive definite:
\\In this case, $r<p+q$ and $F$ is a $(p+q) \times r$ matrix of rank
$r$.
\\Since $ZD^{0.5}$ is a $(p+q) \times (m+n)$ matrix, $F$ is a $(p+q) \times r$
matrix, there exits a matrix $G$ such that $ZD^{0.5}=FG$. This
implies $Z=FGD^{-0.5}$ and $GD^{-0.5}=F^+Z$.
\\$ZLZ^T\gamma=\lambda ZDZ^T\gamma$
\\$\Longrightarrow FGD^{-0.5}LD^{-0.5}G^TF^T\gamma=\lambda FF^T\gamma$
$\Longrightarrow FGD^{-0.5}LD^{-0.5}G^T(F^T\gamma)=\lambda
F(F^T\gamma)$
\\$\Longrightarrow (F^+F)GD^{-0.5}LD^{-0.5}G^T(F^T\gamma)=\lambda (F^T\gamma)$
\\$\Longrightarrow GD^{-0.5}LD^{-0.5}G^T(F^T\gamma)=\lambda
(F^T\gamma)$ $\Longrightarrow F^+ZLZ^T(F^T)^+(F^T\gamma)=\lambda
(F^T\gamma)$
\\$\Longrightarrow$ One solution to $ZLZ^T\gamma = \lambda ZDZ^T\gamma$ is $((F^T)^+x,
\lambda)$, where $x$ and $\lambda$ are eigenvector and eigenvalue
of $F^+ZLZ^T(F^T)^+x=  \lambda x$. Note that eigenvector solution
to Case 2 is not unique.

\begin{theorem}
At level $k$, the multiscale manifold alignment
algorithm achieves the optimal $d_k$ dimensional alignment result
with respect to the cost function $C(\alpha, \beta)$.
\end{theorem}

\textbf{\\Proof:} Let $T=F^+ZLZ^T(F^T)^+$. Since $L$ is positive
semi-definite, $T$ is also positive semi-definite. This means all
eigenvalues of $T \ge 0$, and eigenvectors corresponding to the
smallest non-zero eigenvalues of $T$ are the same as the
eigenvectors corresponding to the largest eigenvalues of $T^+$.
From Theorem 1, we know the solution to generalized eigenvalue
decomposition $ZLZ^T\gamma = \lambda ZDZ^T\gamma$ is given by
$((F^T)^+x, \lambda)$, where $x$ and $\lambda$ are eigenvector and
eigenvalue of $Tx= \lambda x$. Let columns of $P_X$ denote the
eigenvectors corresponding to the $d_k$ largest non-zero
eigenvalues of $T^+$. Then the linear LPP-like solution is
given by  $(F^T)^+P_X$. 

Let columns of $P_Y$ denote $[\phi_k]_{\phi_0}$, the scaling
functions of $T^+$ at level $k$ and $d_k$ be the number of columns
of $[\phi_k]_{\phi_0}$. In our multiscale algorithm, the solution
at level $k$ is provided by $(F^T)^+P_Y$.

From~\citep{dwt}, we know $P_X$ and $P_Y$ span the same space.
This means $P_XP_X^T = P_YP_Y^T$. Since the columns of both $P_X$
and $P_Y$ are orthonormal, we have $P_X^TP_X = P_Y^TP_Y = I$,
where $I$ is an $d_k$$\times$$d_k$ identity matrix. Let $Q =
P_Y^TP_X$, then $P_X = P_XI = P_XP_X^TP_X = P_YP_Y^TP_X =
P_Y(P_Y^TP_X) \Longrightarrow P_X= P_YQ$.

$Q^TQ = QQ^T = I$ and $det(Q^TQ) = (det(Q))^2 = 1$, $det(Q) = 1$.
So $Q$ is a rotation matrix.

Combining the results shown above, the multiscale alignment
algorithm at level $k$ and manifold projections with $d_k$
smallest non-zero eigenvectors achieve the same alignment results
up to a rotation $Q$.\qed

\section{Multiscale Dynamic Time Warping}
\label{wow}

\begin{algorithm}[t]
\caption{Warping on Wavelets (WOW)}
\SetAlgoLined
\KwIn{	X,Y: two time-series data sets \\ d: latent space dimension \\
		$\mu, \tau$: hyper-parameters as described in Algorithm 3.}
\KwOut{	$F^{(X)},F^{(Y)}$: the embeddings of X and Y in the latent space \\
		$W^{(X,Y)}$: the result DTW matrix that provides the alignment of X and Y}
\Begin{
	$t \leftarrow 0$ \\
	$F^{\left(X\right),t}\leftarrow\mbox{MLE}\left(X,\tau \right)$ \\
	$F^{\left(Y\right),t}\leftarrow\mbox{MLE}\left(Y,\tau \right)$ \\
	\Repeat{convergence}{
		$W = \left[ \begin{array}{cc} (1-\mu) W^{(X)} 	& \mu W^{(X,Y),t} 	\\
									  \mu (W^{(X,Y),t})^{T} & (1-\mu) W^{(Y)} \end{array} \right]$ \\
		$\phi^{(Y),t+1},\phi^{(X),t+1}$ $\leftarrow \mbox{MMA}\left(F^{\left(X\right),t},F^{\left(Y\right),t},W,d,\mu, \tau \right)$\\
		$F^{(X),t+1}\leftarrow F^{(X),t}\phi^{(X),t+1}$\\
		$F^{(Y),t+1}\leftarrow F^{(Y),t}\phi^{(Y),t+1}$\\
		$W^{\left(X,Y\right),t+1}\leftarrow\mbox{DTW}\left(F^{\left(X\right),t+1},F^{\left(Y\right),t+1} \right) $ \\
		$t \leftarrow t+1$
	}
	$F^{(X)} \leftarrow F^{(X),t}$; $F^{(Y)} \leftarrow F^{(Y),t}$; $W^{(X,Y)} \leftarrow W^{(X,Y),t}$
}
\end{algorithm}
Algorithm 2 describes a novel multiscale diffusion-wavelet based framework for aligning two sequentially-ordered data sets. 
MLE denotes the multi-scale Laplacian Eigenmaps algorithm described in Figure~\ref{fig:alg}. Also, MMA denotes the multi-scale manifold alignment method described in Section~\ref{mma} as Algorithm 1.  We reformulate the loss function for WOW as:
\begin{equation}
 \begin{array}{l} 
L_{\mbox{WOW}} (\phi^{(X)},\phi^{(Y)},W^{(X,Y)}) \\= ((1-\mu)\displaystyle\sum_{i,j\in X} ||F_{i}^{(X)}\phi^{(X)}-F_{j}^{(X)}\phi^{(X)}||^{2}W_{i,j}^{(X)} \\
+(1-\mu)\displaystyle\sum_{i,j\in Y}||F_{i}^{(Y)}\phi^{(Y)}-F_{j}^{(Y)}\phi^{(Y)}||^{2}W_{i,j}^{(Y)}\\
+\mu \displaystyle\sum_{i\in X,j\in Y}||F_{i}^{(X)}\phi^{(X)}-F_{j}^{(Y)}\phi^{(Y)}||^{2}W_{i,j}^{(X,Y)}
) \end{array}  \end{equation}
which is the same loss function as in linear manifold alignment except that $W^{(X,Y)}$ is now a variable. 

\begin{theorem}
\label{thm3}
Let $L_{\mbox{WOW},t}$ be the loss function $L_{\mbox{WOW}}$ evaluated at \\ $\prod_{i=1}^{t}\phi^{(X),i},\prod_{i=1}^{t}\phi^{(Y),i},W^{(X,Y),t}$ of Algorithm 2.  The sequence $L_{\mbox{WOW},t}$ converges to a minimum as $t \rightarrow \infty$. Therefore, Algorithm 2 will terminate.
\end{theorem}

{\bf Proof:} At any iteration $t$, Algorithm 2 first fixes the correspondence matrix at $W^{(X,Y),t}$.
Now let $L_{\mbox{WOW}}'$ equal $L_{\mbox{WOW}}$ above, except  we replace $F_{i}^{(X)},F_{i}^{(Y)}$ by $F_{i}^{(X),t},F_{i}^{(Y),t}$ and Algorithm 2 minimizes over $\phi^{(X),t+1},\phi^{(Y),t+1}$ using mixed manifold alignment.
Thus,
\begin{equation}
 \begin{array}{l} \label {eqn:linearinequality}
L_{\mbox{WOW}}'(\phi^{(X),t+1},\phi^{(Y),t+1},W^{(X,Y),t}) \\
\leq L_{\mbox{WOW}}'(I,I,W^{(X,Y),t}) \\
= L_{\mbox{WOW}}(\prod_{i=1}^{t}\phi^{(X),i},\prod_{i=1}^{t} \phi^{(Y),i},W^{(X,Y),t}) \\
= L_{\mbox{WOW},t} 
\end{array} 
\end{equation}

since $F^{(X),t}=F^{(X),0}\prod_{i=1}^{t}\phi^{(X),i}$ and $F^{(Y),t}=F^{(Y),0}\prod_{i=1}^{t}\phi^{(X),i}$. We also have:
\begin{equation}
 \begin{array}{l} 
 L_{\mbox{WOW}}'(\phi^{(X),t+1},\phi^{(Y),t+1},W^{(X,Y),t}) \\
 = L_{\mbox{WOW}}(\prod_{i=1}^{t+1}\phi^{(X),i},\prod_{i=1}^{t+1} \phi^{(Y),i},W^{(X,Y),t}) \\
 \leq L_{\mbox{WOW},t} 
\end{array}
\end{equation}

Algorithm 2 then performs DTW to change $W^{(X,Y),t}$ to $W^{(X,Y),t+1}$.
Using the same argument as in the proof of Theorem 2, we have:
\begin{equation}
	\begin{array}{l} 
	L_{\mbox{WOW}}(\prod_{i=1}^{t+1}\phi^{(X),i},\prod_{i=1}^{t+1} \phi^{(Y),i},W^{(X,Y),t+1}) \\
	\leq L_{\mbox{WOW}}(\prod_{i=1}^{t+1}\phi^{(X),i},\prod_{i=1}^{t+1} \phi^{(Y),i},W^{(X,Y),t}) \\
	\leq L_{\mbox{WOW},t}\\
	\Leftrightarrow L_{\mbox{WOW},t+1} \leq L_{\mbox{WOW},t}.
	\end{array}
\end{equation}

\section{Warping on Mixed Manifolds} 
\label{wamm} 
We describe two additional novel variants of dynamic time warping, one called mixed-manifold warping (or WAMM), and the other called curve wrapping. 

\subsection{Low Rank Embedding of Datasets on Mixed Manifolds}
\label{lre}

\begin{figure}
  \centering
  \includegraphics[scale=0.5]{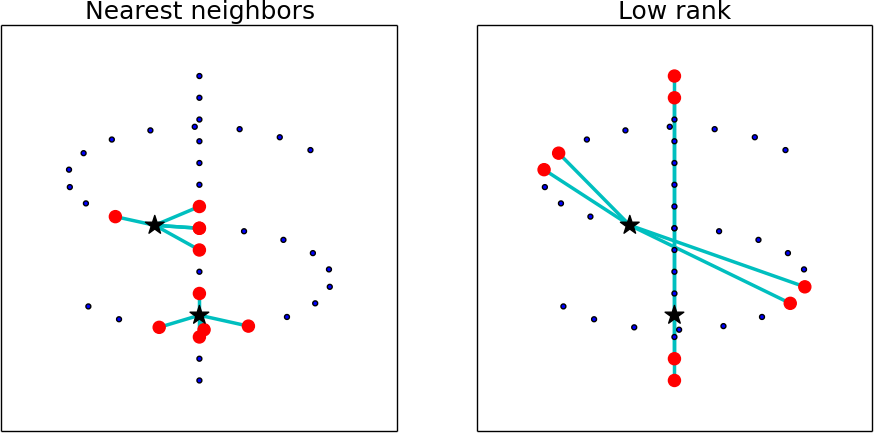}
  \caption{Manifold discovery is nontrivial  when there are multiple intersecting manifolds on which the data potentially lies on.  
  Popular manifold learning algorithms use nearest neighbor construction (on the left), which incorrectly creates \textit{short-circuits},
  whereas a low rank reconstruction (on the right) that correctly differentiates the mixed manifolds.}
  \label{fig:synthetic}
\end{figure}

Traditional manifold learning methods, like LLE \citep{roweis2000lle} and Laplacian eigenmaps \citep{belkin2001laplacian_eigenmaps}, construct a discretized approximation to the underlying manifold by constructing a nearest-neighbor graph of data points in the original high-dimensional space. When data lies on a more complex mixture of manifolds,  methods that rely on nearest neighbor graph construction algorithms  are thus prone to creating spurious inter-manifold connections when mixtures of manifolds are present. These so-called \emph{short-circuit connections}
are most commonly found at junction points between manifolds. Figure~\ref{fig:synthetic} shows an example of this phenomena using a noisy dollar sign data set.

To deal with complex intersecting manifolds, we describe an alternative approach that uses a low-rank reconstruction of the data points that correctly identifies points that lie on mixed manifolds \citep{favaro,DBLP:conf/aaai/BoucherCMD15}. Given a dataset $X$, the first step is to construct a low-rank approximation by reconstructing each point as a linear combination of the other data points. Unlike LLE, which uses a nearest-neighbor approach to manifold construction that is prone to short-circuit errors such as shown in Figure~\ref{fig:synthetic}, our approach is based on a low-rank reconstruction matrix $R$ from minimizing the following objective function:

\begin{algorithm}
\caption{Warping on Mixed Manifolds (WAMM)}
\SetAlgoLined
\KwIn{	X,Y: two time-series data sets \\
		d: latent space dimension \\
		k: number of nearest neighbors used \\
		$\tau$: hyper-parameter for low-rank embedding \\
		$\mu$: hyper-parameter preserving correspondence vs local geometry factor}
\KwOut{	$F^{(X)},F^{(Y)}$: the embeddings of X and Y in the latent space \\
		$W^{(X,Y)}$: the result DTW matrix that provides the alignment of X and Y}
\Begin{
	Set $W^{(X,Y)}_{1,1}=W^{(X,Y)}_{n_{X},n_{Y}}=1$, and $0$ everywhere else
	$t \leftarrow 0$ \\
	\Repeat{convergence}{
		$W = \left[  \begin{array}{cc} (1-\mu) W^{(X)} 	 & \mu W^{(X,Y),t}	 \\
									   \mu (W^{(X,Y),t})^{T} & (1-\mu) W^{(Y)} \end{array} \right]$ \\
		$F^{(X),t+1},F^{(Y),t+1}\leftarrow\mbox{MLE}(F^{(X),t},F^{(Y),t},W,d,\mu, \tau)$\\
		$W^{(X,Y),t+1}\leftarrow\mbox{DTW}(F^{(X),t+1},F^{(Y),t+1})$ \\
		$t \leftarrow t+1$
	}
	$F^{(X)} \leftarrow F^{(X),t}$; $F^{(Y)} \leftarrow F^{(Y),t}$; $W^{(X,Y)} \leftarrow W^{(X,Y),t}$ 
}
\end{algorithm}

In Algorithm 3, $\mbox{MLE(X,Y,W,d,$\mu$)}$ is a function that returns the embedding of $X,Y$ in a $d$ dimensional space using (mixed) manifold alignment with the joint similarity matrix $W$ and parameter $\mu$ described in the previous sections.  To construct such an embedding, we introduce the MME (for mixed-manifold) embedding objective function: 

\begin{equation}
  \label{lra1}
  L_{\mbox{MLE}}(R, \tau) = \min_R \frac{1}{2} \frac{\tau}{2} ||X-XR||_F^2 + ||R||_*,
\end{equation}
where $\lambda>0$, 
$||X||_F=\sqrt{\sum_i \sum_j |x_{i.j}|^2}$ is the Frobenius norm,
and $||X||_*=\sum_i \sigma_i(X)$  is the spectral norm, for singular values $\sigma_i$.

\citep{favaro} prove the following theorem that shows how to minimize the objective function in Equation~\ref{lra1} using a relatively simple SVD computation. 

\begin{theorem}
\label{lr1}
Let $X = U \Sigma V^T$ be the singular value decomposition of a data matrix $X$. Then, the optimal solution to Equation~\ref{lra1} is given by
\begin{equation}
\hat{R} = V_1 (I - \frac{1}{\tau} \Lambda_1^{-2}) V_1^T
\end{equation}
where $U = [U_1 \ U2]$, $\lambda = \mbox{diag}(\Lambda_1 \ \Lambda_2)$, and $V = (V_1 \ V_2)$ are partitioned according to the sets $I_1 = \{i: \lambda_i > \frac{1}{\sqrt{\tau}}\}$, and $I_2 = \{i: \lambda_i \leq \frac{1}{\sqrt{\tau}}\}$.
\end{theorem}

We now describe a slight modification of our previous algorithm, low rank alignment (LRA) \citep{boucher-aaai2015},  to align two general datasets that may lie on a mixture of manifolds. This modification extends LRA in that the latter used a restricted version of MME where the parameter $\tau$ was set to unity. We now assume two data sets $X$ and $Y$ are given, along with the correspondence matrix $C^{(X,Y)}$ describing inter-set correspondences (see equation \ref{wxy}).The goal is to compute a low-dimensional {\em joint} embedding of two datasets $X$ and $Y$, trading off two types of constraints, namely preserving  inter-set correspondences vs.  intra-set geometries. 

The low-rank reconstruction matrices $R^{(X)},R^{(Y)}$ are calculated independently, and can be computed in parallel to reduce compute time. To develop the loss function, we define the block matrices $R,C\in\mathbb{R}^{N\times N}$ as
\begin{equation}
  R = \left[
     \begin{array}{cc}
       R^{(X)} & 0\\
       0 & R^{(Y)}
     \end{array}
   \right]
  \text{ and }
  C = \left[
     \begin{array}{cc}
       0 & C^{(X,Y)}\\
       C^{(Y,X)} & 0
     \end{array}
   \right]
\end{equation}
 and $F\in\mathbb{R}^{N\times d}$ as
\begin{equation}
  F = \left[
     \begin{array}{cc}
       F^{(X)}\\
       F^{(Y)}
     \end{array}
   \right].
\end{equation}

We can write the loss function $L_{\mbox{MMA}}$  for multi-manifold alignment, trading off across-domain correspondence vs. preserving local multi-manifold geometry using a sum of matrix traces:
\begin{align}
  \label{derive_trace}
  L_{\mbox{MMA}}(F, \mu)  &= (1-\mu) tr((F - R F)^\top(F - R F)) \nonumber\\
  & + \mu\sum_{k=1}^d\sum_{i,j=1}^N ||F_{i,k} - F_{j,k}||_2^2 C_{i,j} \nonumber\\
  &= (1-\mu) tr\left(\left((I - R)F\right)^\top (I - R)F\right) \nonumber\\
  & + 2\mu \sum_{k=1}^d F_{\cdot,k}^\top L F_{\cdot,k} \nonumber\\
  &= (1-\mu) tr(F^\top(I - R)^\top(I - R)F) \nonumber\\
  & + 2\mu ~ tr(F^\top L F).
\end{align}

We introduce the constraint $F^\top F=I$ to ensure that the minimization of the loss function $\mathcal{Z}$ is a well-posed problem.
Thus, we have
\begin{align}
  \label{losseqeig}
   L_{\mbox{MMA}}(F, \mu) &  = \argmin_{F:F^\top F=I} (1-\mu) tr(F^\top M F) + 2\mu~ tr(F^\top L F),
\end{align}
where $M = (I - R)^\top(I - R)$.
To construct a loss function from equation (\ref{losseqeig}),
we take the right hand side and introduce the Lagrange multiplier $\Lambda$,
\begin{align}
  \label{lagrange}
  \mathcal{L}(F,\mu, \Lambda) &= (1-\mu) tr(F^\top M F)  + 2 \mu ~ tr(F^\top L F) \nonumber\\
  & + \langle \Lambda,F^\top F-I\rangle.
\end{align}

To minimize equation (\ref{lagrange}), we find the roots of its partial derivatives,
\begin{align}
  \frac{\partial\mathcal{L}}{\partial F} &= 2(1-\mu) M F +  4\mu  L F -  2\Lambda F = 0 \nonumber\\
  \frac{\partial\mathcal{L}}{\partial \Lambda} &= F^\top F - I = 0.
\end{align}
From this system of equations, we are left with the matrix eigenvalue problem
\begin{equation}
  \label{eigprob}
  \left((1-\mu)M + 2 \mu L\right)F = \Lambda F ~~\text{ and }~~ F^\top F = I.
\end{equation}
Therefore, to solve equation (\ref{losseqeig}),
we calculate the $d$ \emph{smallest} non-zero eigenvectors of the matrix
\begin{equation}
  \label{eig_matrix}
  (1-\mu)M + 2 \mu L.
\end{equation}

This eigenvector problem can be solved efficiently because
the matrix $M+L$ is guaranteed to be symmetric, positive semidefinite (PSD), and sparse.
These properties arise from the construction,
\begin{align}
  M + L &=
  \left[
    \begin{array}{cc}
      \left(I-R^{(X)}\right)^2 &  0\\
      0 & \left(I-R^{(Y)}\right)^2
    \end{array}
  \right]\nonumber\\
  &+ \left[
    \begin{array}{cc}
      D^{X} &  -C^{(X,Y)}\\
      \left(-C^{(X,Y)}\right)^\top & D^{Y}
    \end{array}
  \right],
\end{align}
where by construction $D=\left[\begin{array}{cc}D^{X} &  0\\0 & D^{Y}\end{array}\right]$ is a PSD diagonal matrix and $C^{(X,Y)}$ is a sparse matrix.

\subsection{Curve Wrapping}

Curve wrapping is another novel variant that imposes a Laplacian regularization. Since $X$ and $Y$ are points from a time series, we expect $x_{i}, x_{i+1}$ to be close to each other for $1\leq i < n$ and $y_{i}, y_{i+1}$ to be close to each other for $1\leq j < m.$ This leads us to define the following loss function 
\begin{equation}
 \begin{array}{l} 
L_{\mbox{CW}} (F^{(X)},F^{(Y)},W^{(X,Y)}) \\= ((1-\mu)\displaystyle\sum_{i=1}^{n-1} ||F_{i}^{(X)} -F_{i+1}^{(X)}||^{2}W_{i,i+1}^{(X)} \\
+(1-\mu)\displaystyle\sum_{i=1}^{n-1}||F_{i}^{(Y)}-F_{i+1}^{(Y)}||^{2}W_{i,i+1}^{(Y)}\\
+\mu \displaystyle\sum_{i\in X,j\in Y}||F_{i}^{(X)}-F_{j}^{(Y)}||^{2}W_{i,j}^{(X,Y)}
) \end{array}  ,\end{equation}
where we can take $W^X_{i,i+1},W^Y_{i,i+1}=1$ to be either just equal to one or $W^X_{i,i+1}=k^X(x_i,x_{i+1}), W^Y_{i,i+1}=k^Y(y_i,y_{i+1})$ for some appropriate kernel functions $k^X,k^Y.$   Let us define 
\[ W = 
\left[
    \begin{array}{cc}
      (1-\mu)W^X &  \mu W^{(X,Y)}\\
      \mu \left(W^{(X,Y)}\right)^\top & (1-\mu)W^X 
    \end{array}
  \right]
 \]
 and let $L_W$ be the Laplacian corresponding to the adjacency matrix $W$ 
 \[ 
 L_W = \text{diag}(W\cdot 1) - W.
 \]
Let $F = (F_X,F_Y)^T.$ 
We can now express $L_{CW} (F^{(X)},F^{(Y)},W^{(X,Y)}) = F^T L F.$ 
More generally, we expect $x_{i}, x_{i+k}$ to be close to each for all $k\leq k_0,$ where $k_0$ is a small integer. This leads to a slightly different loss function than the above.   

\section{Experimental Results}
\label{sec:experiments} 

\subsection{Synthetic data sets}

\begin{figure} [t]
  \centering
    \includegraphics[height = 5cm, width=0.9\columnwidth,keepaspectratio]{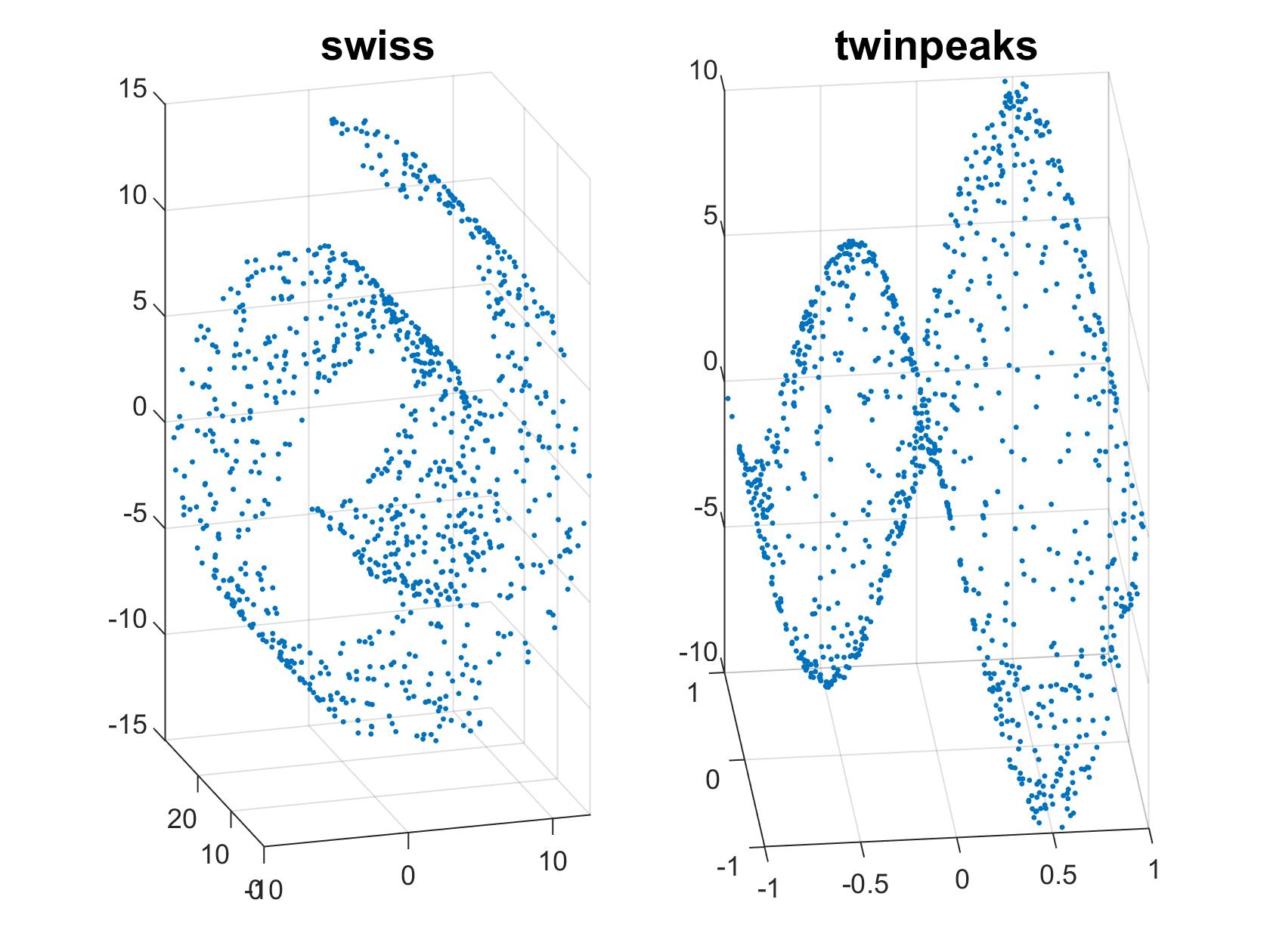}
   \includegraphics[height = 5cm, width=.9\columnwidth,keepaspectratio]{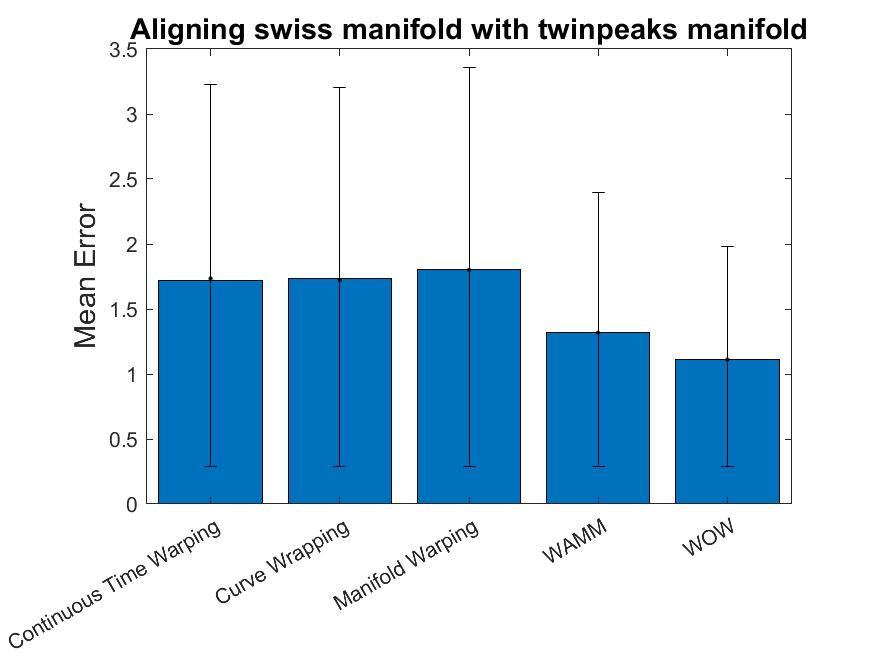}
  \caption{Top: A synthetic  problem  of aligning a swiss roll manifold with a twin peak manifold. The proposed WOW algorithm outperforms both previous methods, such as canonical time warping and manifold warping, as well as two alternative methods called curve wrapping and warping on mixed manifolds. }
  \label{fig:CTWillustration}
\end{figure}

We illustrate the proposed methods with a simple synthetic example  in Figure~\ref{fig:CTWillustration} of aligning two sampled manifolds, a regular swiss roll and a broken swiss roll. In the reprted experiments, alignment error is defined as follows.  Let $p^* = [(1,1), \ldots, (n,n)]$ be the optimal alignment, and let $p = [p1,\ldots, p_I]$ be the alignment output by a particular algorithm. The error$(p,p^*)$between $p$ and $p^*$ is computed by the normalized difference in area under the curve $x=y$ (corresponding to $p^*$) and the piece-wise linear curve obtained by connecting points in $p$. It has the property that
$p \neq p^* \Rightarrow  \mbox{error}(p,p^*) \neq 0$.

\begin{figure} [t]
  \centering
  \includegraphics[height = 4cm, width=0.35\columnwidth,keepaspectratio]{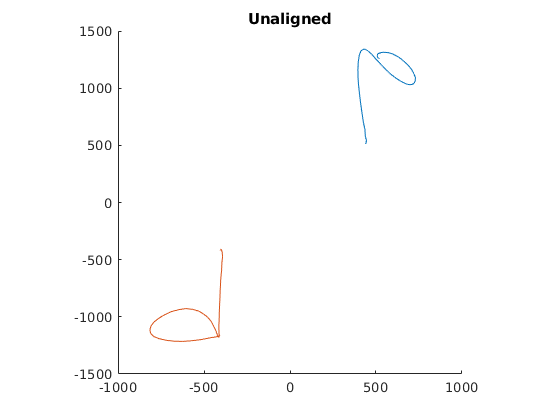}
      \includegraphics[height = 4cm, width=0.35\columnwidth,keepaspectratio]{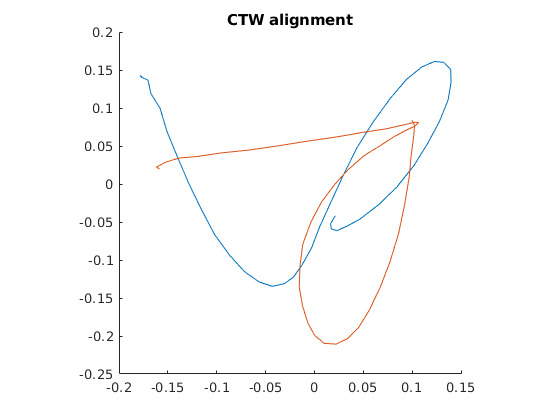}
        \includegraphics[height = 4cm, width=0.55\columnwidth,keepaspectratio]{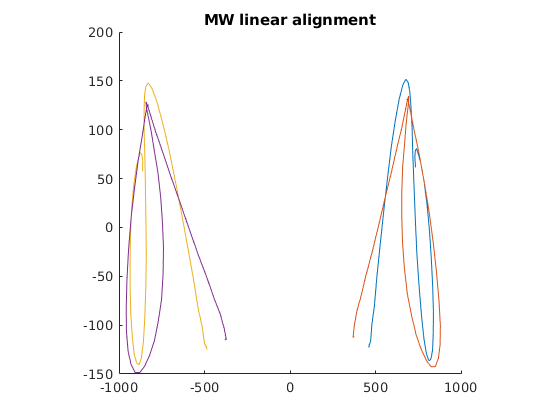}
   \includegraphics[height = 4cm, width=.35\columnwidth,keepaspectratio]{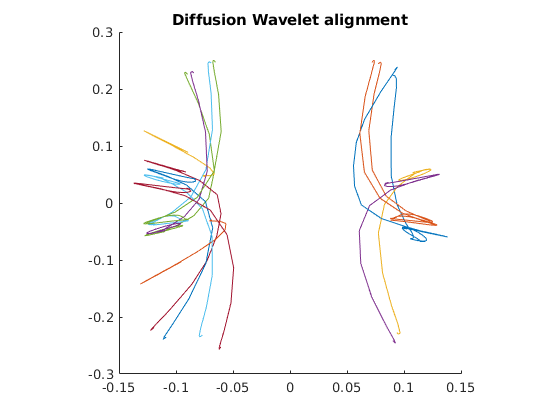}
    \includegraphics[height = 4cm, width=.35\columnwidth,keepaspectratio]{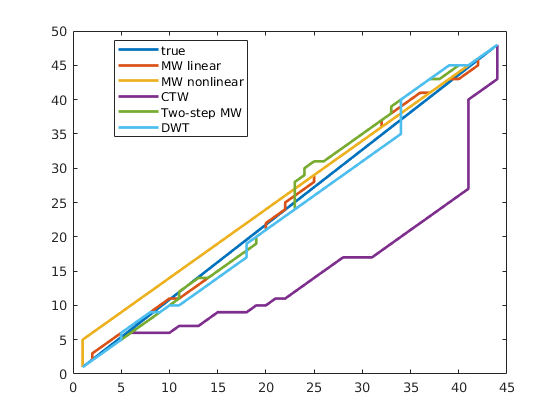}
     \includegraphics[height = 4cm, width=.35\columnwidth,keepaspectratio]{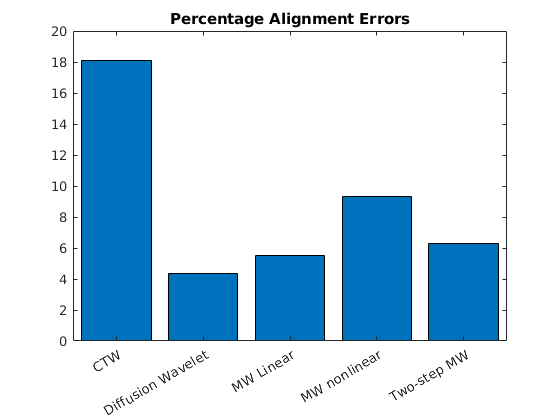}
  \caption{A synthetic problem of aligning rotated digits.}
  \label{fig:CTWillustration2}
\end{figure}

Figure~\ref{fig:CTWillustration2} compares the performance of the proposed WOW algorithm against several other alignment algorithm  on a synthetic rotated digit problem. The original problem is shown on the top left of the panel. The alignments produced by previous methods, such as canonical time-warping  \citep{zhou2009ctw} and manifold (linear, nonlinear, and two-step) warping \citep{hoa-cj-mw}, are compared against the newly proposed WOW algorithm that uses diffusion wavelets. The bottom left plot shows the alignments produced by each method against the ground truth ($45$ degree  line). The bottom right panel computes the alignment error measured in terms of the area difference under each alignment curve vs. the ground truth. 

\subsection{Real World Datasets}

 Table~\ref{tab:algm} summarizes the various proposed novel algorithms and real-world domains used to compare them. The three real-world datasets used to test these algorithms are COIL, the Columbia Object Image Library \citep{coil100}, Human activity recognition (HAR), and the CMU Quality of Life dataset \citep{torre2008motiondata}. 

\begin{table}[h]
\centering
\begin{tabular}{|  c | c |  c  |  c  | }
  \hline
  Method/Domain			& COIL 		& UCI HAR			&   Quality of Life \\
  \hline
 WAMM 	&  	Figure~\ref{fig:coil-exp1}		& 	Figure~\ref{fig:uci-har-expt}	&  Figure~\ref{fig:cmu-expt} \\ \hline
 WOW	& 	Figure~\ref{fig:coil-exp1}	& 	Figure~\ref{fig:uci-har-expt}	&  Figure~\ref{fig:cmu-expt} \\
  \hline
  CW	& 	Figure~\ref{fig:coil-exp1}& 		Figure~\ref{fig:uci-har-expt}	&  Figure~\ref{fig:cmu-expt}	\\
  \hline
Two-step CW		& 	Figure~\ref{fig:coil-exp1}		& 	Figure~\ref{fig:uci-har-expt}			&  Figure~\ref{fig:cmu-expt} \\ \hline
Manifold warping	& 	Figure~\ref{fig:coil-exp1}		& 	Figure~\ref{fig:uci-har-expt}			& Figure~\ref{fig:cmu-expt}  \\

  \hline
\end{tabular}
\caption{Proposed Algorithms and Experimental Domains}
\label{tab:algm}
\end{table}

Table~\ref{tab:hyper} lists the various hyper-parameters used in the above experiments.

\begin{table}[h]
\centering
\begin{tabular}{|  c | c |  c  |  c  | }
  \hline
  ~ 			& COIL 		& UCI HAR			&  CMU Quality of Life \\
  \hline
 $\mu$ 	& 0.5 			& 0.5		& 0.5		\\  \hline
  $\tau$	& 1			& 1	& 1 \\
  \hline
  $d$	& 2	& 2			& 2 	\\
  \hline
$k$			& 10 			& 10			& 10 \\
  \hline
\end{tabular}
\caption{Hyperparameter settings for various datasets}
\label{tab:hyper}
\end{table}

\subsubsection{COIL-100 data set} 

\begin{figure}  [h]
  \centering
   \includegraphics[width=1.0\columnwidth]{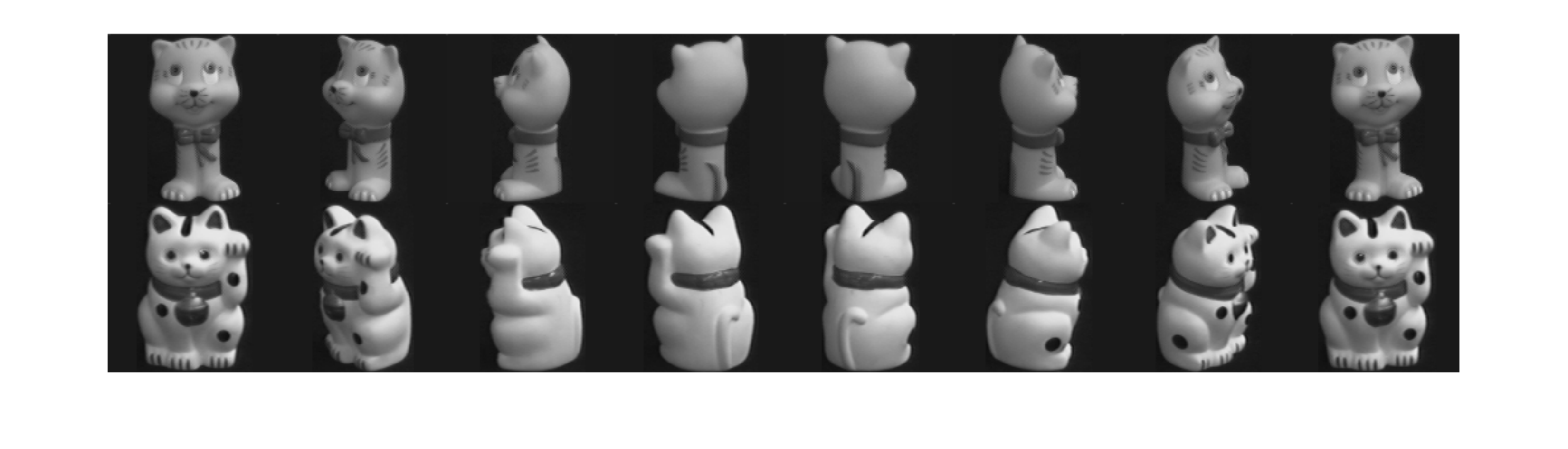}
  \caption{An example of a time-series alignment problem involving rotating objects.}
  \label{fig:coil}
\end{figure}

\begin{figure} [t]
  \centering
    \includegraphics[height = 6cm, width=0.95\columnwidth,keepaspectratio]{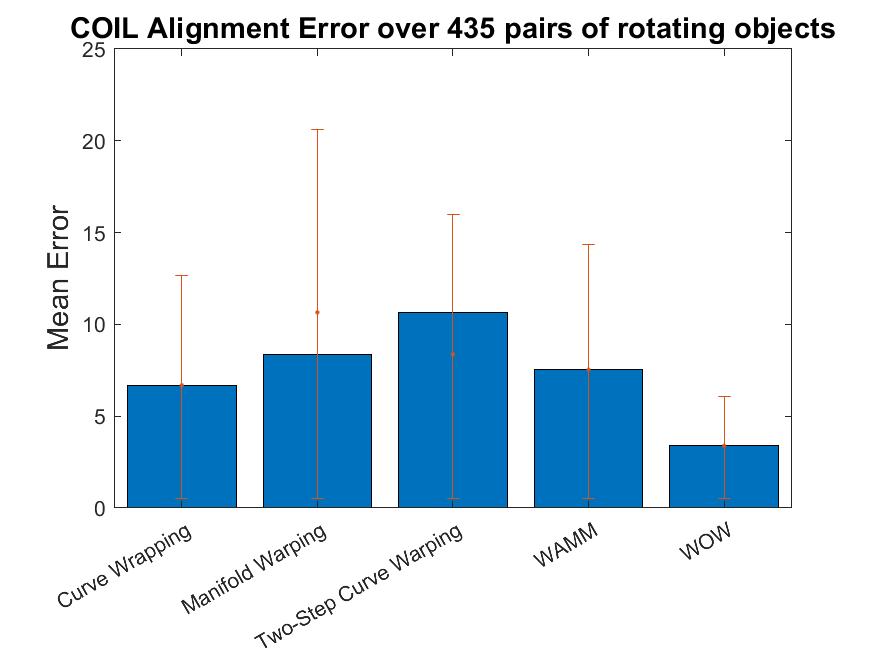}
    \caption{WOW outperforms other alignment algorithms in aligning rotating pairs objects in the COIL vision dataset.}
  \label{fig:coil-exp1}
\end{figure}

The Columbia Object Image Library (COIL100) \citep{coil100} corpus consists of different series of images taken of different objects on a rotating platform (Figure~\ref{fig:coil}). Each series has $72$ images, each $128\times128$ pixels.  Figure~\ref{fig:coil-exp1} reports on experiments over 435 randomly chosen pairs of rotating objects from the COIL dataset, where WOW outperformed the other alignment methods. A paired T-test confirmed the hypothesis that WOW was indeed better to a significance of better than 99\%. 

\subsubsection{Human Activity Recognition} 

\begin{figure}  [h]
  \centering
   \includegraphics[width=0.4\columnwidth]{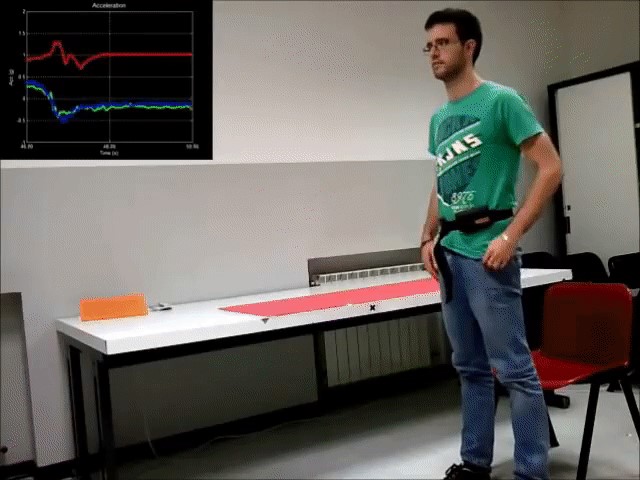}
  \caption{Human Activity Recognition using a Samsung smartphone.}
  \label{fig:har}
\end{figure}

\begin{figure}[p]
  \centering
        \includegraphics[height = 8cm, width=0.95\columnwidth,keepaspectratio]{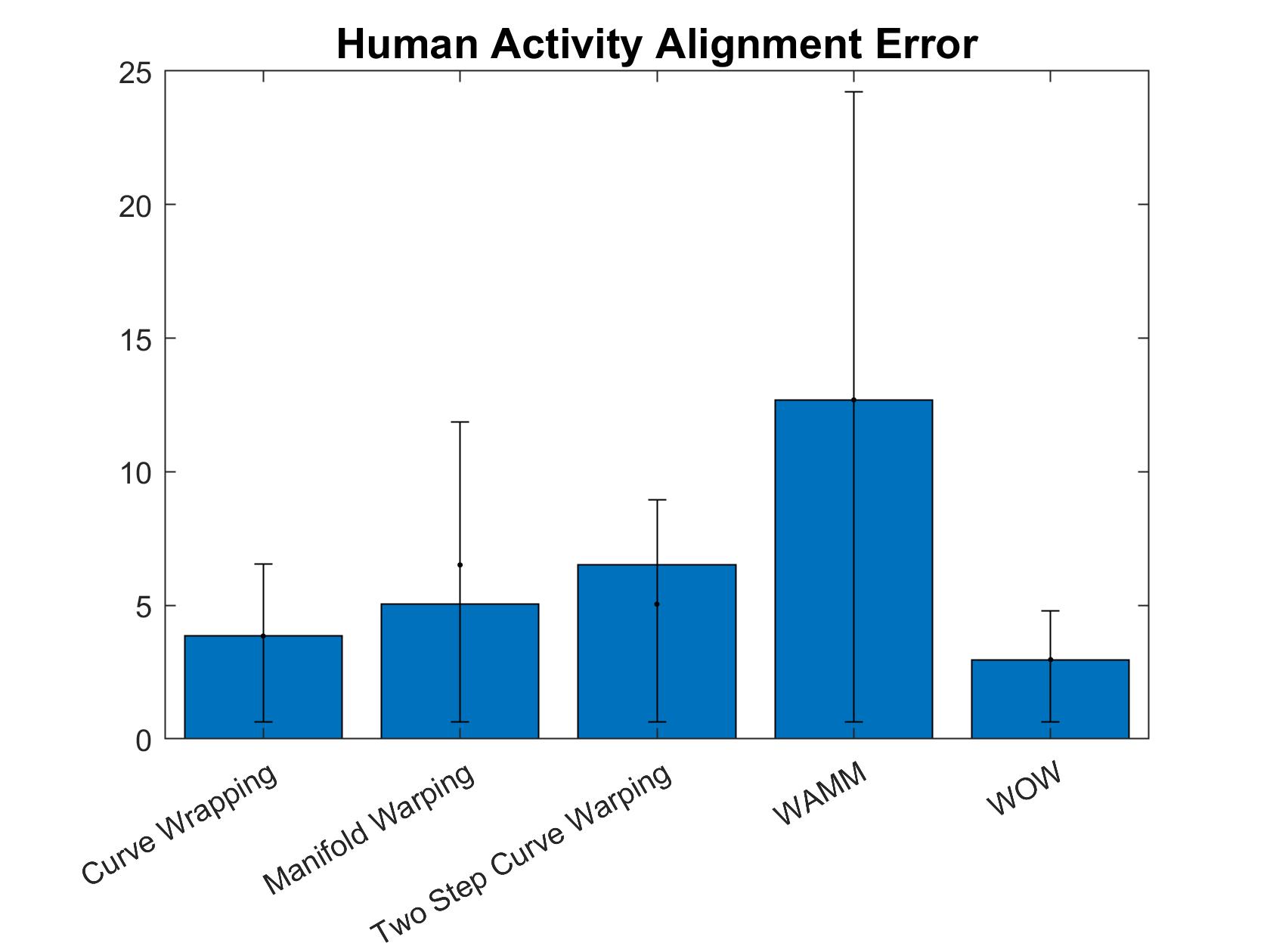}
  \caption{Experimental results on human activity recognition dataset showing mean alignment errors over 100 runs. }
  \label{fig:uci-har-expt}
\end{figure}

The second real-world dataset involves recognition of human activities from recordings made on a Samsung smartphone \citep{uci-har} (see Figure~\ref{fig:har}). \footnote{A video of  this experiment can be found at {\tt https://youtu.be/XOEN9W05\_4A}}. $30$ volunteers performed six 
activities (WALKING, WALKING UPSTAIRS, WALKING DOWNSTAIRS, SITTING,
STANDING, LAYING) while wearing a smartphone (Samsung Galaxy S II) on the
waist. Using its embedded accelerometer and gyroscope, 3-axial linear acceleration and 3-axial angular velocity  measurements were captured at a constant rate of 50Hz. Figure~\ref{fig:uci-har-expt} compares the WOW algorithm against the curve warping, as well as with two varieties of manifold warping.  The results shown are averaged over $100$ trials, where each trial consisted of taking a subject and activity at random, and aligning the $3$-D accelerometer readings with the gyroscope readings. A paired T-test showed the differences between WOW and the other methods were statistically significant at the 95\% or better level.  

\subsubsection{CMU Quality of Life Dataset}

\begin{figure}[p]
  \centering
    \includegraphics[height = 13cm, width=0.85\columnwidth,keepaspectratio]{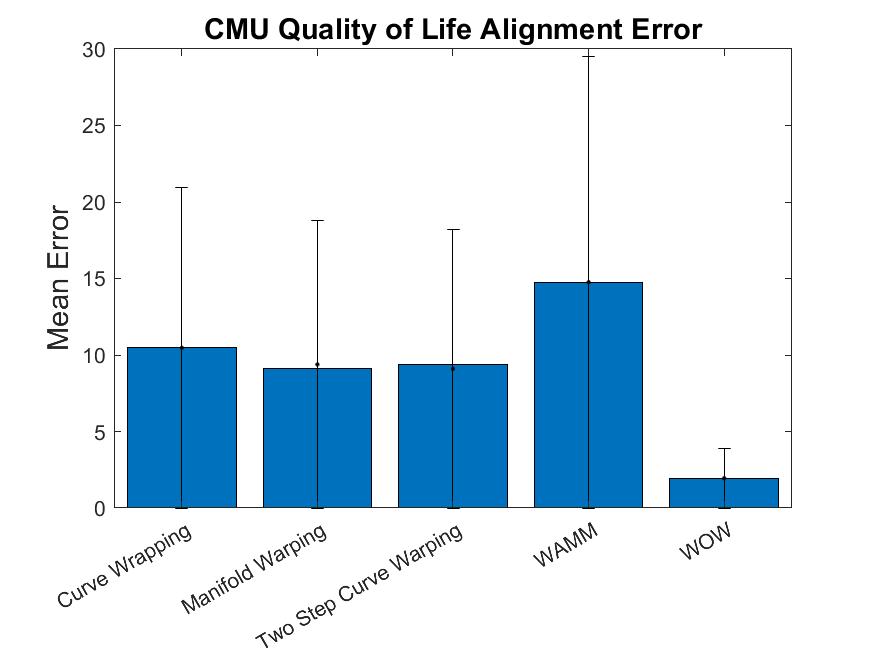}
  \caption{Mean alignment errors on CMU Quality of Life dataset of $25$ randomly chosen segments of $100$ video frames of two subjects making brownies.}
  \label{fig:cmu-expt}
\end{figure}

Our third real-word  experiment uses the kitchen data set \citep{torre2008motiondata} from the CMU Quality of Life Grand Challenge, which records human subjects cooking a variety of dishes (see Figure~\ref{fig:dtw}). The original video frames are NTSC quality (680 x 480), which we subsampled to 60 x 80. We analyzed randomly chosen sequences of 100 frames at various points in two subjects' activities, where the two subjects are both making brownies. As Figure~\ref{fig:cmu-expt} shows, WOW performs significantly better than the other methods, with a paired T-test showing significance better than 99\% with p-values near 0.

\section{Summary and Future Work}
We introduced a novel multiscale time-series alignment framework called WOW, which combines dynamic time warping with diffusion wavelet analysis on graphs. WOW outperforms canonical time warping and manifold warping, two state of the art alignment methods, as well as other novel methods introduced in this paper, such as WAMM and curve wrapping. There are many directions for future work, including exploring faster variants of the proposed algorithms using distributed processors, combining our multiscale algorithms with nonlinear feature extraction methods using deep learning and related techniques, and doing more detailed experimental testing in additional domains.

\end{document}